\pdfoutput=1

\documentclass[11pt]{article}

\usepackage{EACL2023}

\usepackage{times}
\usepackage{latexsym}

\usepackage[T1]{fontenc}

\usepackage[utf8]{inputenc}

\usepackage{microtype}

\usepackage{inconsolata}

\usepackage{graphicx}
\usepackage{comment}
\usepackage{cuted}
\usepackage{amsmath,lipsum}

\usepackage{amsmath,amsfonts,bm}

\def\eqref#1{equation~\ref{#1}}

\def\1{\bm{1}}

\DeclareMathAlphabet{\mathsfit}{\encodingdefault}{\sfdefault}{m}{sl}
\SetMathAlphabet{\mathsfit}{bold}{\encodingdefault}{\sfdefault}{bx}{n}

\usepackage{amsmath}
\usepackage{amssymb}
\usepackage{hyperref}
\usepackage{url}
\usepackage{sidecap}

\usepackage{graphicx}
\usepackage{booktabs}
\usepackage{wrapfig,lipsum,booktabs}
\usepackage{multirow}
\usepackage{booktabs}
\usepackage[normalem]{ulem}
\usepackage[american]{babel}
\usepackage{algorithm}
\usepackage{comment}
\usepackage{algorithmicx}
\usepackage{bm}
\usepackage{enumitem}
\usepackage{amsfonts} 
\usepackage{natbib}
\usepackage{nicefrac}       
\usepackage{microtype}      
\usepackage{subfigure}

\usepackage{xurl}

\title{Distinguishability Calibration to In-Context Learning}

\newcommand{\hq}[1]{\textcolor{black}{#1}}

\newcommand{\yhq}[1]{\textcolor{black}{#1}}

\newcommand*{\affaddr}[1]{#1} 
\newcommand*{\affmark}[1][*]{\textsuperscript{#1}}
\newcommand*{\email}[1]{\texttt{#1}}

\begin{document}

\author{%
Hongjing Li\affmark[1]\footnotemark[1], Hanqi Yan\affmark[1]\footnotemark[1], Yanran Li, Li Qian\affmark[2], Yulan He\affmark[1,3,4], and Lin Gui\affmark[3]\\
\affaddr{\affmark[1]Department of Computer Science, University of Warwick, UK}\\
\affaddr{\affmark[2]Xiaomi AI Lab, China}\\
\affaddr{\affmark[3]Department of informatics, King's College London, UK}\\
\affaddr{\affmark[4]The Alan Turing Institute, UK}\\
\email{\{Hongjing.Li,Hanqi.Yan\}@warwick.ac.uk, yanranli.summer@gmail.com,}\\
\email{qianli@xiaomi.com, \{yulan.he,lin.1.gui\}@kcl.ac.uk}
}

\maketitle

\renewcommand{\thefootnote}{\fnsymbol{footnote}} 
\footnotetext[1]{Equal contribution.} 
\renewcommand{\thefootnote}{\arabic{footnote}}

\begin{abstract}
Recent years have witnessed increasing interests in prompt-based learning in which models can be trained on only a few annotated instances, making them suitable in low-resource settings. When using prompt-based learning for text classification, the goal is to use a pre-trained language model (PLM) to predict a missing token in a pre-defined template given an input text, which can be mapped to a class label. 
However, PLMs built on the transformer architecture tend to generate similar output embeddings,  
\hq{making it difficult to discriminate between different class labels.} The problem is further exacerbated when dealing with classification tasks involving many fine-grained class labels. In this work, we alleviate this \emph{information diffusion} issue, i.e., different tokens share a large proportion of similar information after going through stacked multiple self-attention layers in a transformer, by proposing a calibration method built on feature transformations through rotation and scaling to map a PLM-encoded embedding 
into a new metric space 
to guarantee the distinguishability of the resulting embeddings. 
Furthermore, we take the advantage of hyperbolic embeddings to capture the hierarchical relations among fine-grained class-associated token embedding 
by a coarse-to-fine metric learning strategy to enhance the distinguishability of the learned output embeddings. 
Extensive experiments on the three datasets under various settings demonstrate the effectiveness of our approach. \footnote{Our code can be found at \url{https://github.com/donttal/TARA}}
\end{abstract}

\section{Introduction}
Large pre-trained language models (PLMs)~\citep{DBLP:conf/naacl/DevlinCLT19,DBLP:conf/iclr/LanCGGSS20,liu2019roberta} have been achieved state-of-the-art performance in many Natural Language Processing (NLP) downstream tasks. More recently, the PLMs with prompt learning demonstrate surprising capabilities in numerous tasks both in NLP and  computer vision, even outperforming their fine-tuned counterparts~\cite{DBLP:conf/nips/BrownMRSKDNSSAA20,DBLP:journals/corr/abs-2107-13586,DBLP:conf/emnlp/LesterAC21,zhou2022learning,DBLP:journals/corr/abs-2110-04544}. 

\begingroup
\setlength{\tabcolsep}{6pt} 
\renewcommand{\arraystretch}{1} 

\begin{table}[ht]
\small
\begin{tabular}{ll}
\toprule[1pt]
\textbf{Train\#1}: & Gotta protect’em! It was [MASK].                               \\
\textbf{Train\#2}: & That's why it's only 20\$. It was [MASK].                             \\  \hline
\textbf{Test}:    & On a boat trip to Denmark. It was [MASK]. \\ \bottomrule[1pt]
\end{tabular}
\caption{The prompt templates for emotion classification. The samples are from \textsl{GoEmotion}~\cite{DBLP:conf/acl/DemszkyMKCNR20} dataset.}
\label{tab:example}
\vspace{-0.5cm}
\end{table}

In an emotion classification task shown in Table \ref{tab:example}, an input sentence $X$, followed by a prompt, \emph{``It was [MASK]''}, is fed to a PLM to predict the missing token at the position of \textsl{[MASK]}. The predicted word can be used to identify the emotion label of the input sentence. Such few-shot learning generates a probability distribution over the \textsl{[MASK]} conditioning on the given prompt/context, which is considered as in-context learning of language models. 

However, as in-context learning does not require updating PLM parameters, there arises the problem of distribution mismatch between the data used for LM pre-training and the test samples used in in-context learning, which hinders the full exploitation of the knowledge encoded in PLMs~\cite{DBLP:conf/iclr/XieRL022,DBLP:conf/icml/ZhaoWFK021,DBLP:journals/corr/abs-2202-06687,DBLP:conf/naacl/ShinLAKKKCLPHS22}. To alleviate the context shift, existing methods rely on prior knowledge to increase the overlapping between the two distributions. For example, \textsl{PTR}~\citep{han2021ptr} appends domain-agnostic tokens to prompts to discriminate the domains, such as \emph{``sports''}, \emph{``politics''}. Another line of studies designs sophisticated handcrafted verbalizers to map the test samples onto the label word space derived from PLMs~\cite{DBLP:conf/eacl/SchickS21,DBLP:conf/acl/GaoFC20}. Although the gradient-optimized verbalizers~\citep{DBLP:conf/acl/HuDWLWLWS22} are proposed to ease the human effort and can be adapted to different downstream tasks via training, it is still considered inferior to the manual verbalizers, especially in both the few-shot and zero-shot settings where training data are scarce. 

In this paper, we first show that PLMs have an inherent \emph{information diffusion} issue in their generated output token embeddings, which share a large proportion of similar information after going through a stack of transformer layers~\cite{DBLP:conf/iclr/GaoHTQWL19,pmlr-v180-yan22b}. Such token embeddings occupy a narrow cone, leading to largely overlapped output distributions when applied to in-context learning. Next, we elaborate that the overlapped output distributions would violate the distinguishability condition~\cite{DBLP:conf/iclr/XieRL022} under in-context learning. 
To this end, we propose to flatten the singular value distributions of the output embeddings generated from PLMs to shape the space spanned by the singular values to a desirable manifold. 
On the one hand, we apply an orthogonal and a scaling constraints to the weight matrix applied to the output embeddings, which can avoid exploding and vanishing values in the feature matrix~\cite{DBLP:journals/corr/SaxeMG13}, leading to better  discriminative features when trained with limited labelled data. On the other hand, we leverage hyperbolic embeddings to capture the hierarchical relations among fine-grained class labels of training examples to further enhance the distinguishability of output embeddings.

Our proposed framework has been implemented on top of existing prompt-based few-shot learning methods and it demonstrates an average {5.86\%} performance improvement of F1-measure on three classification tasks under 100-shot learning. We also verify that the improvement stems from a more balanced singular value distribution for the output features and the learnt hierarchical feature space.

In summary, our contributions include: 
\begin{itemize}
    \item We propose a transformation-based constraint to output embeddings by rotation and ratio balancing which is able to guarantee the distinguishability of learned embeddings. 
    \item The proposed hyperbolic embedding-based metric learning strategy not only improves the performance of prompt learning but also measures the relation between different categories. 
    \item  The experimental results outperform many strong baselines and the visualisation illustrates that the proposed method is able to project the embedding to a less overlapping distribution and improve the interpretability and distinguishability of output. Specifically, across three evaluated datasets, our method surpasses the state-of-the-art by 9.60\%, 5.11\% and 2.87\%, respectively, in the 100-shot setting.
\end{itemize} 

\section{Related works}
\paragraph{\emph{Information diffusion} in PLMs.} In a typical $L$-layer transformer-based PLM, assuming 
the prompt is a concatenation of a few training examples and a test input $X_\text{test}$, consisting of $m$ tokens in total, the goal of in-context learning is to predict the output distribution over the masked token at the $t$-th position, \textsl{[MASK]}. It is formally defined by the following equation:
\begin{gather*}
    p(\mathcal{O}_t|X_{\text{test}}) = \mathbb{E}_{h\sim p_\text{prompt}(h|X_{\text{test}})}[p(\mathcal{O}_t|X_{\text{test}},h,\theta)],
\label{eqn:self-attention}
\end{gather*}
\noindent where $h$ denotes the last-layer hidden state corresponding to the token of $X_{\text{test}}$, $\theta$ is the parameters in prompt-based learning. 

Although we have limited knowledge of the output distribution $p(\mathcal{O}_t|X_{\text{test}})$ over token \textsl{[MASK]}, many existing studies analyzed the geometry properties of the last layer feature $h^{L}$, and examined its effects in downstream tasks~\cite{DBLP:conf/icml/GoyalCRCSV20,DBLP:conf/acl/ZhouS22}. Due to the softmax bottleneck~\cite{DBLP:conf/iclr/YangDSC18} and the likelihood loss in language generation tasks~\cite{DBLP:conf/iclr/GaoHTQWL19}, the output feature distribution in PLMs tends to be anisotropic and rank-deficient, which limits the expressiveness of the generated representations.~\citet{DBLP:conf/icml/GoyalCRCSV20} discussed the information diffusion issue among tokens within a sentence that feeding the tokens in different positions for classification only resulted in a 1.2\% variance in classification accuracy. 
\citet{DBLP:conf/iclr/GaoHTQWL19} explored the information diffusion among different sentences via singular value decomposing and they found that the singular value distributions are skewed especially in deeper PLM layers, i.e., larger singular values become more predominant compared to the smaller ones.

\begin{figure*}[!t]
    \centering
    \includegraphics[width=0.85\textwidth,trim={0 0 0 0},clip]{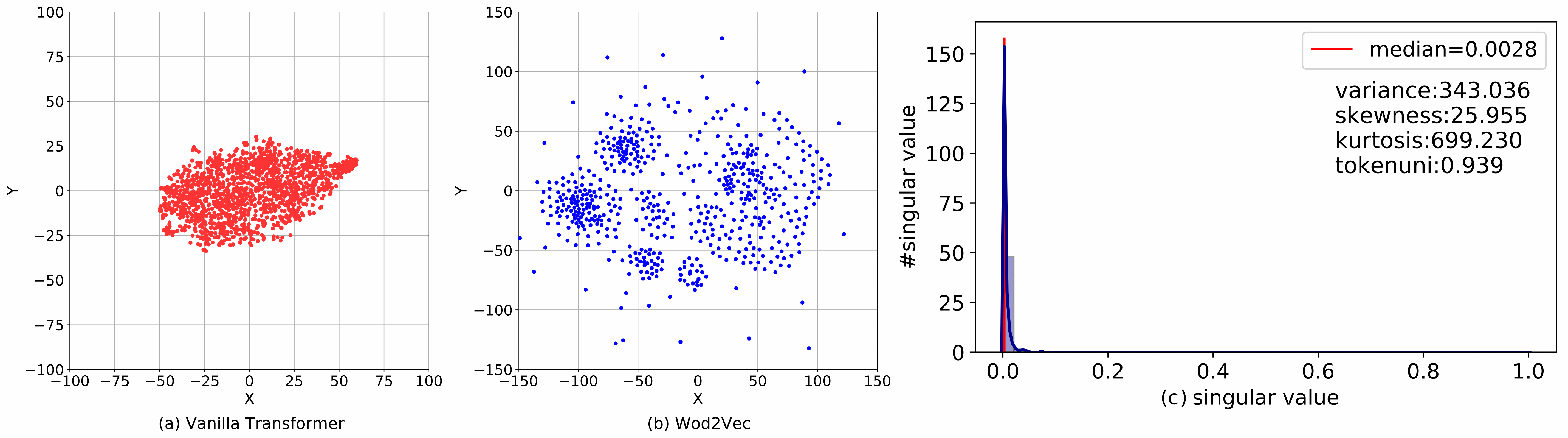}
\caption{\textbf{(a)}: The mapping results of \yhq{1,500} \textsl{[MASK]} tokens randomly sampled from the \textsl{GoEmotions} dataset. Each red dot is the output representations derived from prompt-based learning for the \textsl{[MASK]} token of an input example, which will be used to predict the masked token in the corresponding position. \textbf{(b)}: Each blue dot is the static word representation of the corresponding predicted token with the largest probability on \textsl{[MASK]} for one of the 1,500 samples in (a) from the \textsl{GoEmotions} dataset. \textbf{(c)}: Singular value distribution (after normalisation) of the output representations of the randomly selected \yhq{1,500} \textsl{[MASK]}s. It is clear that the representations are dominated by very few singular values.}
    \label{fig:narrowcone_sv}
\vspace{-0.3cm}
\end{figure*}

\paragraph{Context shift in in-context learning.} Many researchers studied the distribution shift (aka. domain shift) between the pretraining corpora and test samples and proposed solutions to decrease the performance variance in prompt-based few-shot learning~\citep{DBLP:conf/iclr/XieRL022,DBLP:conf/icml/ZhaoWFK021,DBLP:conf/acl/HuDWLWLWS22,zhou2022learning,DBLP:conf/naacl/ShinLAKKKCLPHS22}. On the one hand, some in-context learning methods incorporated domain-specific words or learnable tokens in the prompt to discriminate different context. 
~\citet{ben2022pada} proposed to first generate the name of the domain and then generate domain-related features (DRFs) conditioned on the domain in a supervised manner. Both the generated domain name and DRFs were used as the prompt fed to the model. On the other hand, the sophisticated verbalizers contributed to minimising the distance between the two distributions~\cite{schick-etal-2020-automatically,schick-schutze-2021-just,DBLP:conf/acl/GaoFC20,DBLP:conf/acl/HuDWLWLWS22}. To broaden the coverage of single-choice verbalizer, \textsl{Knowledge Prompt Tuning (KPT)}~\cite{DBLP:conf/acl/HuDWLWLWS22} used the knowledge graph to extract more topic-related words as label words and then refine the label word candidates. 
To incorporate prior knowledge to calibrate the context shift, ~\citet{DBLP:conf/iclr/XieRL022} 
simplified a language model as the Hidden Markov Model, where the observed tokens are sampled from a family of concepts 
and proposed the \emph{distinguishability condition} to measure context shift as the Kullback–Leibler (KL) divergence.

\section{Contextual Calibration for Output Distribution}
\label{sec:outputdistribution}
Many existing methods calibrate the probabilities of the generated tokens in a language model in order to improve the generation quality. In prompt-based learning, we want to find out if the output distribution $p(\mathcal{O}_t|X_{\text{test}})$ or the output feature $h^{\text{[mask]}}$, which is a part of the hidden representation from the last layer of a PLM, $h^{\ell}$,  
suffers from the \emph{information diffusion} issue and occupies a narrow cone. 
We take RoBERTa-based prompt learning as an example and derive the value of $h^{\text{[mask]}}$ from \yhq{1,500} randomly selected test samples from an emotion classification dataset, \textsl{GoEmotions}~\cite{DBLP:conf/acl/DemszkyMKCNR20}, and visualise the results in a 2D plane in Figure~\ref{fig:narrowcone_sv}(a). For comparison, we select the predicted token with the largest probability on each \textsl{[MASK]} and map their corresponding vectors from Word2Vec~\cite{DBLP:journals/corr/abs-1301-3781} to a 2D plane in \ref{fig:narrowcone_sv}(b).
It is clear that the word embeddings learned from Word2Vec has a more uniform distribution around the origin. In contrast, the representations derived by RoBERTa degenerate into a narrow cone, which implies limited expressiveness. Inspired by the approach proposed in \cite{pmlr-v180-yan22b}, we display the singular value distribution of $h^{\text{[mask]}}$ and calculate the distribution statistics, i.e., the matrix moment and the average cosine similarity between every \textsl{[MASK]} pair in Figure~\ref{fig:narrowcone_sv}(c). From the empirical results, we can see that the value of the hidden representation for \textsl{[MASK]} in different samples share much similar information with the token uniformity value ~\cite{pmlr-v180-yan22b} (\emph{tokenuni} in Figure~\ref{fig:narrowcone_sv}(c)) of 0.939. This shows that most $h^{\text{[mask]}}$ concentrates at very few singular values, \hq{which implies a severe information diffusion issue.}

\subsection{Uniform Ratio-based Distinguishability}
\label{sec:uniform}

Although many calibration methods have been proposed, few of them focuses on explicitly addressing the information diffusion issue in the prompt-based learning framework. \hq{One main challenge in this task is that the unlabelled data used in language model pre-training is significantly larger than the labelled samples used for prompt tuning. Hence, the optimised distribution in prompt-based few-shot learning can be very different from the true distribution.} To avoid inheriting the \emph{information issue} caused in the pre-training phase, we propose a calibration method to reduce the skewness of the output token distributions, 
such that the output representations are evenly distributed in the embedding space. The idea is to rotate the original embedding space to an isotropic metric space by an inner product-based operator on a learnable basis. For each dimension of the basis, we use the inner product to measure its relevance with a given input. The dimension-dependent relevance scores are sent to a Multi-layer Perceptron (MLP) decoder to generate the calibrated output embedding for final prediction.

The framework of the proposed calibration method is shown in Figure \ref{fig:framework}. In practice, due to the small size of training samples in prompt learning, the relevance scores might be dominated by very few dimensions. Therefore, inspired by \citet{https://doi.org/10.48550/arxiv.2208.05134}, who proposed a ratio estimator to balance the distribution from different label categories, we design a scaling matrix in our isotropic distribution scenario. That is, for both labelled and unlabelled data, the multi-class ratio between different dimensions should be similar.
\begin{figure}[h]
\centering
\includegraphics[trim={0 5 10 2},clip,width=0.35\textwidth]{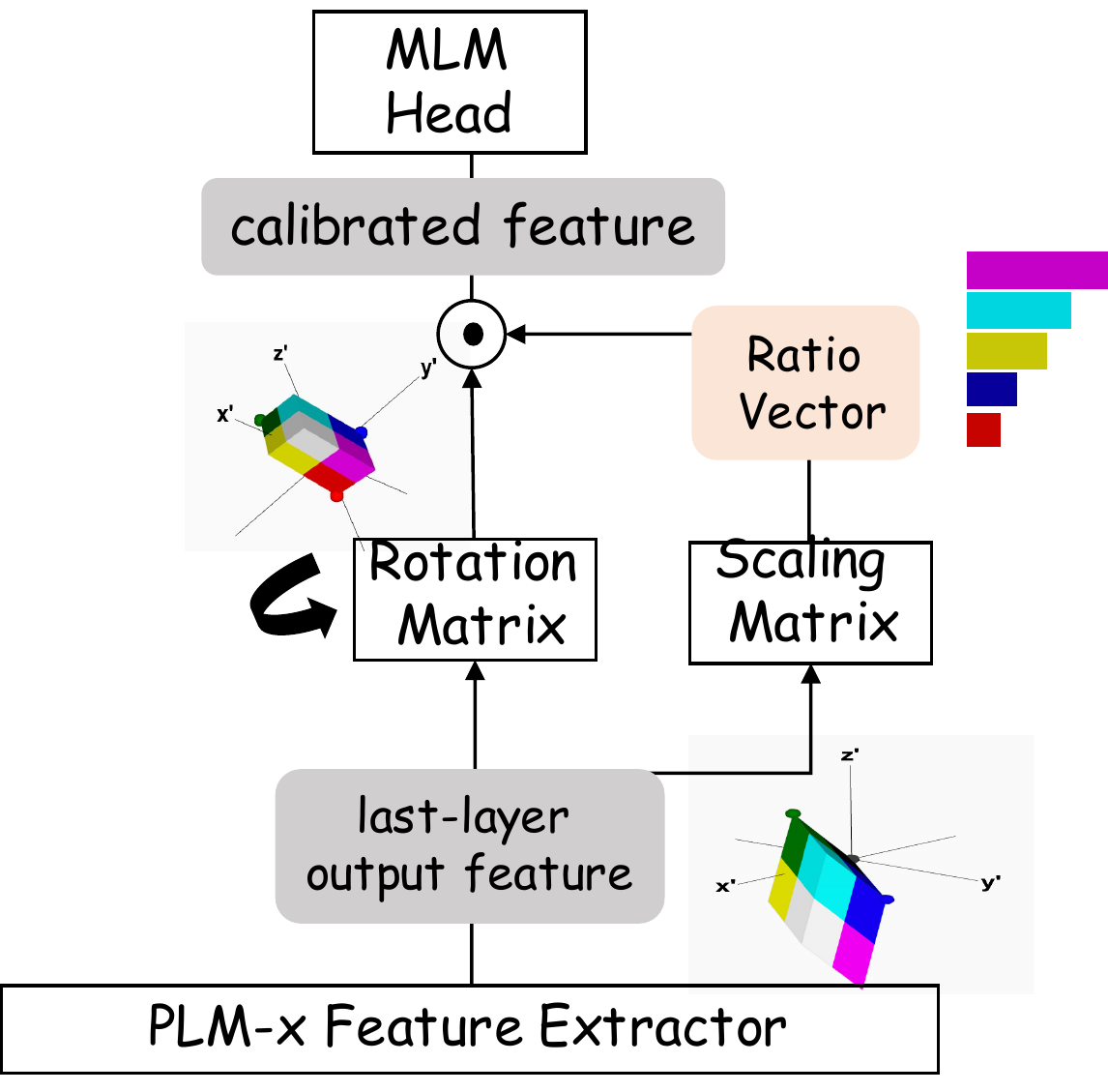}
\caption{Our proposed calibration method is applied to the output embeddings from the last layer of a PLM. After being transformed with a rotation matrix through a Multi-layer Perception (MLP), the resulting output feature is assumed to have a more balanced singular value distribution in different basis directions. Moreover, as the vector norm on each projected direction would change in the new base, we derive a ratio vector to balance the distribution along the rotated directions.}
\label{fig:framework}
\vspace{-0.2cm}
\end{figure}

Concretely, assuming we have $N$ labelled data $\{y_j, x_j \}_{j=1}^N$ and $M$ unlabelled data from pre-training $\{ x_j\}_{j=N+1}^{N+M}$, where $x_j$ is the input sample, $y_j$ is the true label, and $M \gg N$. To simplify the notation, in the rest of this paper, we use $x_j$ to represent the feature of the last embedding layer and $h_j$ to represent the output of our calibrated feature. Then, for the representation of a masked token, $x_j$, we assume there are $K$ isotropic directions in the metric space and the corresponding inner product based relevance score is:
\begin{equation}
    \mathcal{H}_k(x_j) = \sigma(\langle x_j, W_k \rangle), \, (1 \leq k \leq K),
    \label{eq:rot}
\end{equation}
\noindent where $\sigma(\cdot)$ is the softmax activation function. Here, we can define a \textbf{rotation matrix} based on $W_k$ 
since Eq. (\ref{eq:rot}) projects an input embedding onto a new metric space by rotation. To guarantee the orthogonality of the basis in the new metric space, we use the following regulariser during training:


\begin{equation}
    \mathcal{L}_{orth} = \left\|W^{\top} W-\bm{I}\right\|_{2}^{2},
    \label{eq:orth}
\end{equation}

\noindent where $W$ is the stacking of $\{W_k\}_{k=1}^K$. Correspondingly, for each dimension $k$, we can define a ratio score which aims to better separate them 
to avoid the skewed distribution by minimising the following loss:
\begin{equation}
   \mathcal{L}_t = \frac{1}{N+M} \Sigma_{k=1}^K \Sigma_{j=1}^{N+M}|| \mathcal{R}_k(x_j) - \frac{1}{K}||^2,
   \label{eq:unsuper}
\end{equation}

\noindent where $\mathcal{R}_k(x_j)$ is an MLP-based estimator with a softmax activation:
\begin{equation}
   \mathcal{R}_k(x_j) = \sigma(S_k \cdot x_j + \beta).
\label{eq:lt}
\end{equation}

By minimising $\mathcal{L}_t$, even if one input sample $x_j$ is similar to a basis vector along a popular dimension $k$, there will still be a probability to assign it a low ratio score $\mathcal{R}_k(x_j)$ if there are other samples which are more closer to the basis vector in dimension $k$. In this way, we can balance the distribution after rotation. We define the stacking of $S_k$ as a \textbf{scaling matrix} which aims to distribute $x_j$ uniformly into $K$ clusters in the metric space.\footnote{We measured the impact of different weight initialisations on $S_k$ in Appendix~\ref{sec:weight_initial}.} 

However, it is difficult to optimise the loss defined in Eq. (\ref{eq:unsuper}) since the size of the unlabelled data for pre-training is much larger than the labelled data and the unlabelled data is usually unseen to the downstream tasks. We instead define an alternative optimisation objective. 
First, according to Eq. (\ref{eq:unsuper}), we need to ensure that for any two dimensions $k$ and $t$, we have    $\frac{1}{N+M} e^{S_k \cdot x_j} = \frac{1}{N+M} e^{S_t \cdot x_j}$. By the Jensen's inequality, we have the following lower bound: $e^{\frac{1}{N+M}{S_k \cdot x_j}}\leq \frac{1}{N+M} e^{S_k \cdot x_j}$, in which we can achieve the lower bound for any two independent dimensions by taking $\frac{1}{N+M}{S_k \cdot x_j} = \frac{1}{N+M}{S_t \cdot x_j} $. It means that for any two dimensions, the sum of their ratio scores should be similar. As such, Eq. (\ref{eq:unsuper}) can be approximated by:
\begin{equation}
    \mathcal{L}_t \sim \Sigma_{k=1}^K (||S_k||^2 - 1 )^2.
\label{eq:lt2}
\end{equation}

Accordingly, we can define the distinguishability loss in a more general form by both the relevance score and the ratio score without the need of sampling from unlabelled data:  
\begin{equation}
    \mathcal{L}_{dis} = \mathcal{L}_{orth} + \mathcal{L}_t.
    \label{eq:dis}
\end{equation}

From our findings in Section~\ref{sec:outputdistribution}, much information encoded by the output representations generated by the last layer of a PLM only occupies a space spanned by very few singular value directions. This leads to the information diffusion issue. 
Therefore, our solution here is to re-project the output features into a new  hyperplane, in which the information is more evenly distributed in different directions, and at the same time we can derive a ratio vector by aggregating the rotated components.

\subsection{Supervised Prompt Learning}
\label{sec:TARA}

By our proposed distinguishability loss-based learning in Section 
\ref{sec:uniform}, an input embedding has been separated into vectors along $K$ independent dimensions. Then, for the labelled data $\{ x_j \}_{j=1}^N$, we propose 
to use $k$ independent decoders to produce the final prediction. The decoding 
result is based on the relevance score and ratio score on each independent 
dimension: 
\begin{equation}
    \bm{h}_{j} =  \Sigma_{i=k}^K {\rm Decoder}_k(\mathcal{H}_k(x_j) \cdot \mathcal{R}_k(x_j)),
\end{equation}

\noindent where the ${\rm Decoder}_k$ is a decoder for the $k$-th dimension. Then the representation of $\bm{h}_{j}$ can be used in the \yhq{verbalizer $p_{verbalizer}(\hat{\mathcal{O}}|\bm{h}_{j})$}, where $\hat{\mathcal{O}}$ is the predicted masked token. Finally, the cross-entropy loss $H$ is defined by the predicted $\hat{\mathcal{O}}$ and the true label $y_j$:
\begin{equation}
    \mathcal{L}_{cls}(x_j) =  H(y_j, p_{verbalizer}(\hat{\mathcal{O}}|\bm{h}_{j})).
    \label{eq:Lcls}
\end{equation}

By combining the uniform ratio-based distinguishability loss of $\mathcal{L}_{dis}$ and the prompt-based classification loss $\mathcal{L}_{cls}$, we propose our first model, named as \textbf{T}ransformation based \textbf{A}daptation for \textbf{R}atio b\textbf{A}lanced 
 (\textbf{\textbf{TARA}}) prompt learning,  which aims to minimise $\mathcal{L}_{\textbf{TARA}} =\mathcal{L}_{cls}(x_j)+\mathcal{L}_{dis}$. Note that $\mathcal{L}_{cls}(x_j)$ is the default loss term in all the baselines and our proposed methods.

\subsection{Dimension Rotation by Hyperbolic Embeddings}
\label{hyperbolic}

In Section \ref{sec:uniform}, we project the input mask embedding into a $K$ dimensional metric space to avoid skewed distributions. However, we ignore the potential class relations between the dimensions. For example, in emotion classification, both the emotions of `gratitude' and `approval' belong to the \emph{coarse} positive class, but they are associated with different \emph{fine-grained} labels in the GoEmotions dataset~\cite{DBLP:conf/acl/DemszkyMKCNR20}. Hence, in this section, we only consider those positive pairs under the same coarse category to achieve a better class disambiguation by a proxy based metric learning \cite{DBLP:conf/iccv/Movshovitz-Attias17,DBLP:conf/wacv/YangBZGS22}, which uses an anchor vector to represent a category for metric loss optimisation and capture the hierarchical structure between coarse- and fine-grained labels in the hyperbolic space.

\noindent\textbf{Strategies for Constructing Sample Pairs}. 
Inspired by the hierarchical structure of coarse-to-fine emotion categories, we assume that a fine-grained emotion should be close to the coarse-grained emotion it belongs to. 
To implement this idea, we construct sample-anchor pairs $(\bm{h}_{j}, z_{i}^{+})$ 
for training, where $\bm{h}_{j}$ is the representation for prompt prediction and $z_{i}^{+} \in \mathbb{R}^{d}$ is a learnable anchor representation for each coarse class. %

\noindent\textbf{Metric Learning in a Hyperbolic Space}. To maximise the similarity in sample-anchor positive pairs, where the sample and the anchor share the same coarse-grained label, while minimising the similarity in negative pairs, we adopt the following metric learning objective: 
\begin{equation}
 \mathcal{L}_{metric}(\bm{h}_{j}) = - log\frac{e^{-d(\bm{h}_{j},z_{pj}^{+} )}}{\sum_{i=1}^{C} e^{-d(\bm{h}_{j},z_{i}^{+})}} \label{metric learning loss},
 \end{equation}
where $ \{(\bm{h}_{j}, z_{i}^{+})\}_{i=1}^C$ represents a set of sample-anchor pairs that we constructed for each sample $i$, $C$ denotes the number of anchors, $z_{pj}^{+}$ is the representation of positive pairing anchor of $j$-th sample, and $d(\cdot)$ is the hyperbolic distance metric defined by the Poincar\'e ball model of the hyperbolic space \cite{DBLP:journals/corr/NickelK17}. In a $n$-dimensional hyperbolic space, all points will fall into a unit open interval: $\mathcal{I}^{n} =\left \{ x \in \mathbf{R^{n}}| \left \| x \right \| < 1 \right \}$, where $\left \| \cdot \right \| $ donates the Euclidean norm. The distance $d(\cdot)$ between two points $u,v \in \mathcal{I}^{n}$ can be formulated as:
 \begin{equation}
 \label{distance}
 d(u,v) = \operatorname{arcosh}(1+2\frac{\left \| u-v \right \| ^2}{(1-\left \| u \right \|^2)(1-\left \| v \right \|^2)} ). 
 \end{equation}

The motivation of using $\mathcal{L}_{metric}(\bm{h}_{j})$ is to push similar categories together in the metric space. Hence, we can obtain our final learningn objective 
by adding the loss of tree-structured metric learning $\mathcal{L}_{metric}(\bm{h}_{j})$ to \textbf{TARA} as: 
\begin{equation}
 \label{eq:total_loss}
 \mathcal{L}_{final} = \mathcal{L}_{cls}(x_j)  + \mathcal{L}_{metric}(\bm{h}_{j}) + \mathcal{L}_{dis}.
\end{equation}

For a comparison, we propose a variant called \textbf{TML} by keeping the learning architectue and simply adding $ \mathcal{L}_{metric}(\bm{h}_{j})$ to the classification loss of $\mathcal{L}_{cls}(x_j)$, but without the ratio balancing term of $\mathcal{L}_{dis}$, that is, $\mathcal{L}_{\textbf{TML}} = \mathcal{L}_{cls}(x_j)  + \mathcal{L}_{metric}(\bm{h}_{j})$.

\section{Experiments}

\paragraph{Datasets}

We evaluate our proposed approach on three multi-class text classification datasets,  the \textsl{Emotion}\footnote{https://huggingface.co/datasets/emotion}~\citep{saravia-etal-2018-carer} dataset, an academic paper classification dataset, \textsl{WOS}~\citep{kowsari2017HDLTex}, and a fine-grained emotion classification dataset, \textsl{GoEmotions}\footnote{https://huggingface.co/datasets/go\_emotions}~\citep{DBLP:conf/acl/DemszkyMKCNR20}. All of these datasets have hierarchical label structures. The datasets statistics are shown in Table~\ref{tab:dataset}.

\begingroup
\setlength{\tabcolsep}{6pt} 
\renewcommand{\arraystretch}{1} 

\begin{table}[!h]
\small
\centering
\resizebox{0.45\textwidth}{!}{%
\begin{tabular}{lrrrr}
\toprule[1pt]
\textbf{Name} & \textbf{\#Classes} & \textbf{\#Train} & \textbf{\#Dev} & \textbf{\#Test} \\ \midrule
\textsl{\textbf{Emotion}}              & 6          & 16,000      & 2,000        & 2,000   \\ 
\textsl{\textbf{WOS}} & 11          & 5,736      & 1,147        & 1,147 \\
\textsl{\textbf{GoEmotions}}           & 28         & 23,485       & 2,956      & 2,984     \\ 
\bottomrule[1pt]
\end{tabular}}
\caption{Dataset statistics.}
\label{tab:dataset}
\end{table}

\noindent For all datasets, we remove punctuation, digits, and special characters that do not have specific semantic meanings. For the \textsl{Emotion} dataset which consists of tweet, we also remove user mentions.  

\paragraph{Baselines}

We implement our proposed framework on top of the commonly used prompt-based learning methods and compare it with existing approaches including those which can be used for learning more discriminative representations: 
\begin{itemize}
    \item \textsl{Prompt-baselines}. Three commonly used prompt-based methods are selected including \textsl{Soft Prompts}~\citep{DBLP:conf/nips/BrownMRSKDNSSAA20}, \textsl{Prompt-Tuning}~\citep{DBLP:conf/emnlp/LesterAC21} and \textsl{PTR}~\citep{han2021ptr}. The best-performing methods is used as the default prompt-based training method for the following three comparison models, and denoted as \textsl{Prompt-baseline}.
    \footnote{The detailed performance of these three prompt-based training methods is shown in Table ~\ref{tab:template_res}. We use \textsl{PTR} for \textsl{GoEmotion}, and use \textsl{P-tuning} for the other two datasets.}
    \item \textsl{KPT}~\citep{DBLP:conf/acl/HuDWLWLWS22}. It uses a knowledge graph to incorporate topic-related label words to increase the coverage of the verbaliser.
    \item \textsl{Context Calibration}~\citep{DBLP:conf/icml/ZhaoWFK021}. This method calibrates the output representations by one-layer linear transformation, whose weight matrix is optimised to be diagonal. 
    \item \textsl{Proxy-NCA}~\citep{DBLP:conf/iccv/Movshovitz-Attias17}. It creates a proxy for each class and uses the Neighbourhood Component Analysis (NCA) loss to pull samples closer to their assigned proxies while pushing negative samples away.
\end{itemize}

\paragraph{Prompt Settings}
As the performance of prompt-based methods heavily relies on prompt templates and verbalisers, we use the same template and verbaliser for all models for fair comparison. 
The prompt templates are shown in Table~\ref{tab:templates}. 
The original class labels are used as label words in the verbaliser as in~\cite{DBLP:conf/eacl/SchickS21}.

\begingroup
\setlength{\tabcolsep}{6pt} 
\renewcommand{\arraystretch}{1} 

\begin{table}[ht]
\centering
\small
\resizebox{0.98\linewidth}{!}{
\begin{tabular}{p{1.4cm}p{6.2cm}}
\toprule[1pt]
\textbf{Datasets}   & \textbf{Prompt template}                                      \\ \midrule
\textsl{\textbf{Emotion}}& \emph{\textless{X}\textgreater It's [MASK].}                                  \\ 
\textsl{\textbf{WOS}}&\emph{\textless{X}\textgreater  The domain of the text is [MASK].}            \\ 
\textsl{\textbf{GoEmotions}} &  \emph{\textless{X}\textgreater The emotional aspect of this text is [MASK].} \\ 
\bottomrule[1pt]
\end{tabular}}
\caption{Prompt templates used in three datasets.}
\label{tab:templates}
\vspace{-0.3cm}
\end{table}

\begin{table*}[htbp]
\small
\centering
\resizebox{0.98\textwidth}{!}{%
\begin{tabular}{l|cccc|cccc|cccc}
\toprule[1pt]
\multicolumn{1}{l}{}                                              & \multicolumn{4}{c|}{\textbf{Emotion}} & \multicolumn{4}{c|}{\textbf{WOS}}   & \multicolumn{4}{c}{\textbf{GoEmotions}}       \\ \midrule[0.5pt]
\textbf{$K$-shot} & \textbf{5}      & \textbf{10}    & \textbf{50}    & \textbf{100}   & \textbf{5}     & \textbf{10}    & \textbf{50}    & \textbf{100}   & \textbf{5}     & \textbf{10}    & \textbf{50}    & \textbf{100}   \\ \midrule[0.5pt]
\textsl{\textbf{Prompt-baseline}}                                            & 0.336 & 0.363 & 0.431 & 0.625 & 0.236 & 0.252 & 0.359 & 0.435 & 0.161  & 0.173 & 0.281 & 0.310 \\ 
\textsl{\textbf{Proxy-NCA}}                                                         & 0.333 & 0.384 & 0.412 & 0.637 & 0.214 & 0.246 & 0.295 & 0.383 &  0.149  & 0.166 & 0.208 & 0.233 \\ 
\textsl{\textbf{Context Calibration}}  & 0.337 & 0.352 & 0.531 & 0.706 & 0.212 & 0.361 & 0.687 & 0.707 & 0.164  & 0.224 & 0.355 & 0.420\\ 
\hline
\textbf{\textbf{TML}}                                                                & 0.339 & 0.387 & 0.466 & 0.699 & 0.229 & 0.277 & 0.372 & 0.529 & 0.158  & 0.227 & 0.309 & 0.355 \\ 
\textbf{\textbf{TARA}}                                                               & 0.348 & 0.401 & 0.697 & 0.783 & 0.245 & 0.418 & 0.705 & 0.728 & 0.172  & 0.249 & 0.364 & 0.442\\ 
\textbf{Ours full model}                                                           & \textbf{0.355} & \textbf{0.441} & \textbf{0.713} & \textbf{0.802} & \textbf{0.278} & \textbf{0.439} & \textbf{0.719} & \textbf{0.757} & \textbf{0.206}  & \textbf{0.255} & \textbf{0.384} & \textbf{0.448}\\ 
\bottomrule[1pt]
\end{tabular}
}
\caption{Weighted F1 scores on three Datasets. \hq{The proposed \textbf{TML} is better than \textsl{Proxy-NCA}. Our full method (\textbf{TML}+\textbf{TARA}) achieves the best performance among all the settings.}}
\label{tab:balanced_results}
\vspace{-0.3cm}
\end{table*}

\subsection{Few-shot Learning on Three Datasets}
We randomly select $k$ different training samples for few-shot learning and show the results across the three datasets in Table~\ref{tab:balanced_results}.

\underline{For metric-learning}, \hq{\textsl{Proxy-NCA} with contrastive loss leads to performance degradation compared to the \textsl{Prompt-baseline}, with more significant performance drops on the \textsl{GoEmotions} dataset, which has the largest label categories. By contrast, \textbf{TML} gives better results over the \textsl{Prompt-baseline} and \textsl{Proxy-NCA}, showing its efficiency in encoding hierarchical relations between the coarse- and fine-grained labels. It can be further demonstrated in Figure~\ref{heatmap}, which shows the similarity matrix ($28\times28$) of the 28 fine-grained emotion labels from 3 high-level categories, i.e., \emph{``anger''},\emph{``joy''} and\emph{``sad''}. The results of \textsl{Proxy-NCA} in (c) are similar to the \textsl{Prompt-baseline} as shown in (b). Our proposed \textbf{TML} in (d) can capture the hierarchical relations among the 
28 labels, where the correlations among labels belonging to the same high-level emotion category are similar. By comparison, we replace the hyperbolic distance in \textbf{TML} with the Euclidean distance and show the results in (c). It can be observed that the resulting label embeddings fail to exhibit different patterns within and across different high-level emotion categories.}

\hq{\underline{For the calibration methods}, \textsl{Context Calibration} and \textbf{TARA} are overall better than the \textsl{Prompt-baseline}. This shows that the simple linear transformation of the output representations can greatly improve the performance of prompt-based learning. The superior performance of \textbf{TARA} over \textsl{Context Calibration} demonstrates the benefit of using our proposed rotation and scaling transformations. Combining \textbf{TML} with \textbf{TARA}, our full model achieves the best performance and the improvements are more predominant when $K$ is larger.} In the 100-shot setting, our method surpasses the state-of-the-art method, \textsl{Context Calibration}, by 9.6\% on Emotion, 5.1\% on WOS, and 2.9\% on GoEmotions, respectively, verifying its superiority in the few-shot text classification task.

\begin{figure}[t!]
    \centering
    \includegraphics[width=0.75\linewidth]{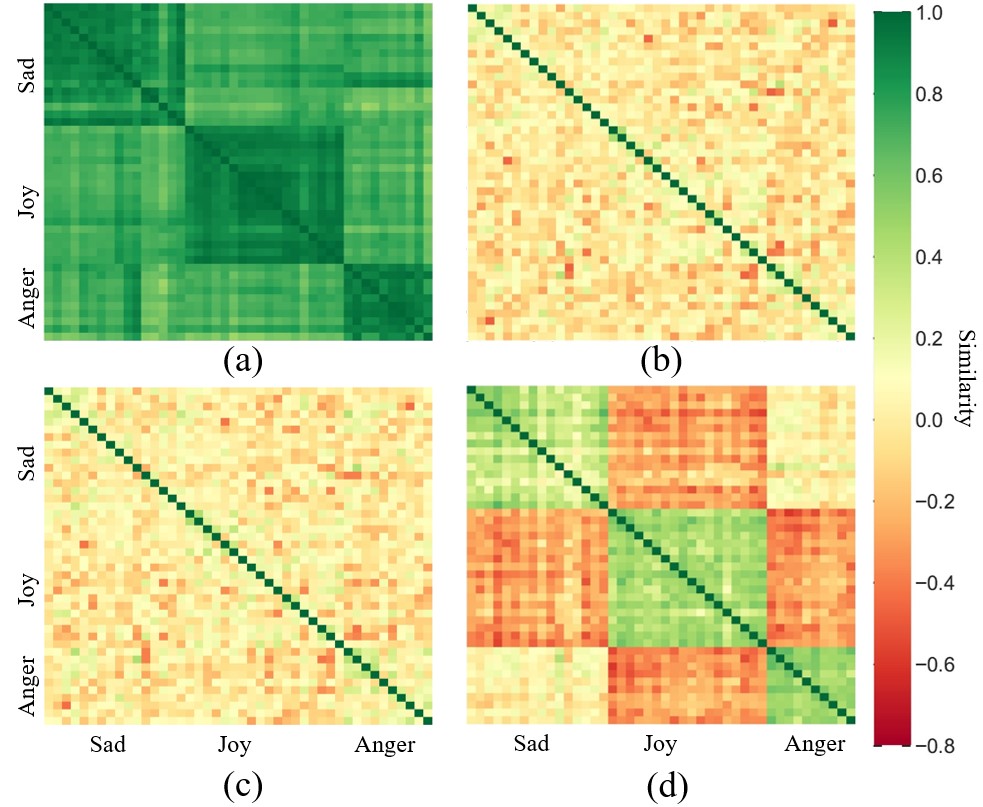}
    \caption{Heatmap for the pair-wised cosine similarity of fine-grained classes on \textsl{GoEmotion}. (a) Label representations from PLM without fine-tuning. (b) Fine-tuned label representations by classification module only. (c) Fine-tuned label representations with proposed constraint but based on Euclidean distance, i.e., \textsl{Proxy-NCA}. (d) Fine-tuned label representations by \textbf{TML}.}
    \label{heatmap}
\vspace{-0.50cm}
\end{figure}

\begin{figure*}[h]
\centering
\subfigure[\textsl{Emotion}.]{
\begin{minipage}[t]{0.48\linewidth}
\centering
\includegraphics[width=\textwidth,trim={0 0 0 0},clip]{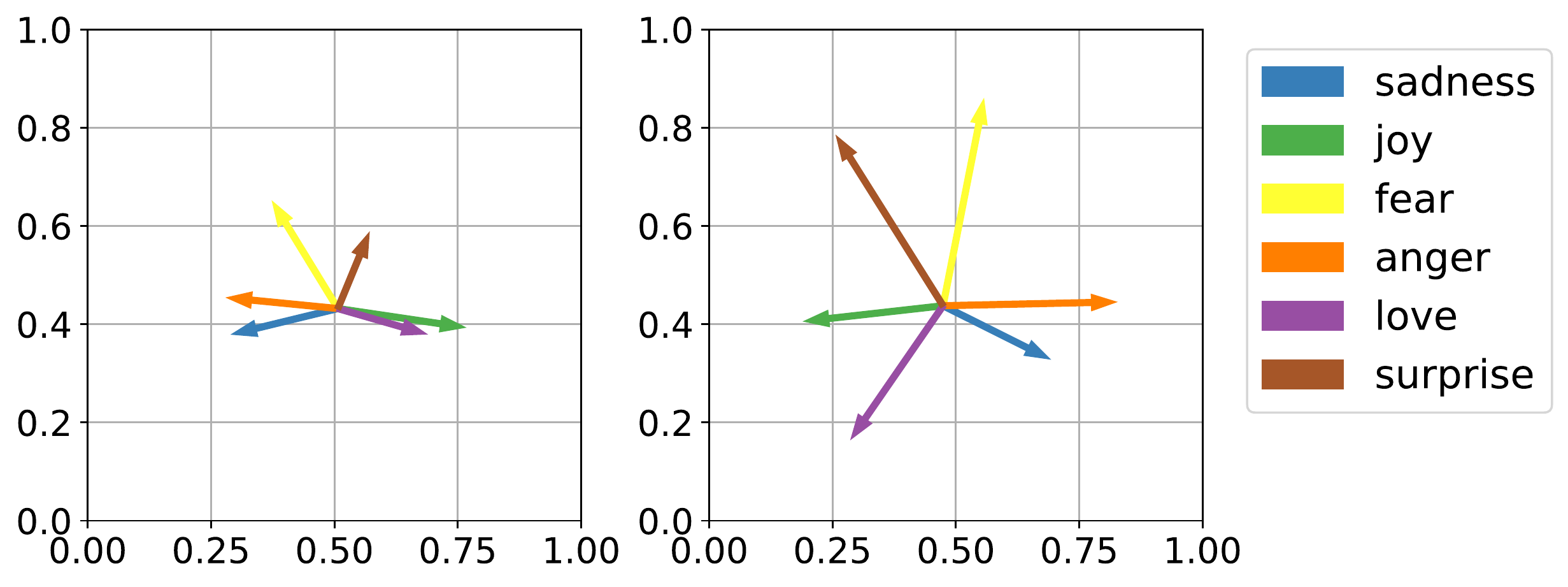}
\end{minipage}%
}%
\subfigure[\textsl{WOS}.]{
\begin{minipage}[t]{0.48\linewidth}
\centering
\includegraphics[width=\textwidth,trim={0 0 0 0},clip]{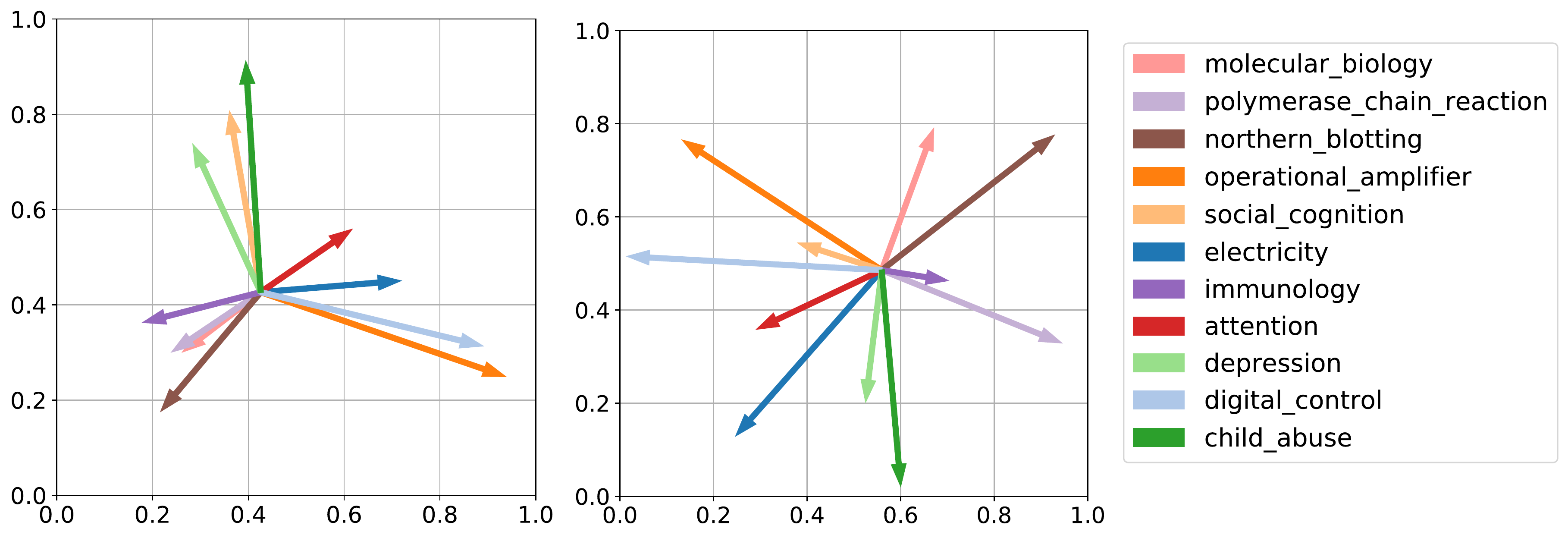}
\end{minipage}%
}
\\
\subfigure[\textsl{GoEmotion}.]{
\begin{minipage}[t]{0.85\linewidth}
\centering
\includegraphics[width=\textwidth,trim={0 0 0 0},clip]{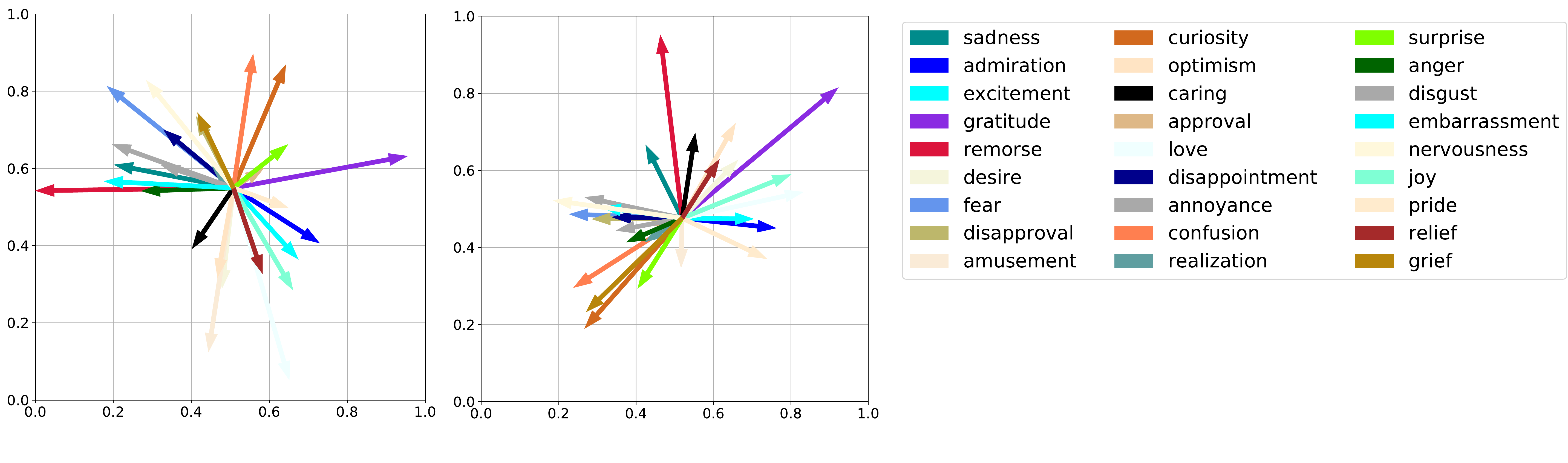}
\end{minipage}%
}
\caption{The PCA projection of the output representations belonging to different classes. In each sub-figure, the \textbf{left figure is the prompt-baseline}, while \textbf{the right figure is our method}. It is clear that our method distributes the output representations more evenly in the embedding space, while the output representations from the baseline appear to be more concentrated.}
\label{fig:pca}
\vspace{-0.5cm}
\end{figure*}

\subsection{Information Diffusion Alleviation} 

In addition to the classification results, we also examine the characteristics of the generated output representations to check whether the information diffusion issue has been addressed. Figure~\ref{fig:pca} shows the PCA projection results of all the \textsl{[MASK]} representations, i.e., $h^{\text{[MASK]}}$ in the test samples, which are colour-coded according to their assigned class labels by the model. It is clear that our method can generate more widely distributed [MASK] representations, therefore better reducing the overlaps of the features from different class labels. For example, in the \textsl{Emotion} dataset, the output features from the baseline model mostly reside along the horizontal direction, while ours distribute more evenly across different directions.\footnote{The T-SNE results and singular value distribution of the output representations in \textsl{Emotion} and \textsl{GoEmotions} are shown in Figure~\ref{app:tsne_fig} and Figure~\ref{fig:sv_dis}.}

We also calculate the summary statistics of the singular value distribution of the output features, as well as the average similarity between every two \textsl{[MASK]} pairs. 
The results are shown in Table~\ref{tab:dist_sta}.
\hq{The average cosine similarity (\textsl{CosSim}) between every token pair is used as a proxy measure of the degree of information diffusion. We can observe that the \textsl{CosSim} value calculated on the output representations generated by our model is significantly lower compared to the other baselines. We also observe an increase in the median and the decrease in variance of the singular value distribution from our model outputs in comparison to the prompt learning baseline. 
The results show that our model produces the output representations which have a more balanced singular value distribution. 
The smaller skewness value further verifies that our proposed model can generate isotropic representations where the embedding dimensions are uncorrelated. }

\begin{table}[t]
\centering
\tabcolsep=2pt
\small
\resizebox{0.98\linewidth}{!}{
\begin{tabular}{lcccc}
\toprule[1pt]
\multicolumn{1}{l}{}     & \multicolumn{1}{l}{\textbf{Median}} & \multicolumn{1}{l}{\textbf{Variance}} & \multicolumn{1}{l}{\textbf{Skewness}} & \multicolumn{1}{l}{\textbf{CosSim}}      \\ \midrule
\textsl{\textbf{Emotion-prompt}}      & 0.0028                     & 371.9                        & 24.57                        & 0.898                         \\ 
\textsl{\textbf{Emotion-Ours}}    & 0.0145                     & 5.211                        & 8.960                        & 0.256                         \\  \midrule[0.5pt]
\textsl{\textbf{WOS-prompt}}          & 0.0036                     & 235.8                        & 22.06                        & 0.817                         \\ 
\textsl{\textbf{WOS-Ours}}      & 0.0117                     & 5.681                        & 9.088                        & 0.191                        \\ \midrule[0.5pt]
\textsl{\textbf{GoEmotions-prompt}}   & 0.0028                     & 822.1                        & 24.64                       & 0.899                         \\
\textsl{\textbf{GoEmotions-Ours}} & 0.0268                     & 11.20                       & 7.728                        & 0.243                         \\ 
\bottomrule[1pt]
\end{tabular}}
\caption{The statistics of the singular value distribution of the output features, as well as the average cosine similarity of all \textsl{[MASK]} token pairs.}
\label{tab:dist_sta}
\end{table}

\begin{table}[t]
\centering
\tabcolsep=3pt
\small
\resizebox{0.45\textwidth}{!}{%
\begin{tabular}{llcccc}
\toprule[1pt]
           & \textbf{Ours}                      & \multicolumn{1}{l}{\textbf{w/o $\mathcal{L}_{orth}$}} & \multicolumn{1}{l}{\textbf{w/o $\mathcal{L}_t$}} & \multicolumn{1}{l}{\textbf{w/o $l_2$}} & \multicolumn{1}{l}{\textbf{w/o all}} \\ \midrule
\textbf{Emotion}    & \multicolumn{1}{c}{0.802} & 0.725                      & 0.719                      & 0.723                      & 0.724                                                                                 \\
\textbf{WOS}        & 0.757                     & 0.728                      & 0.687                      & 0.741                      & 0.699  \\ 
\textbf{GoEmotions} & 0.448                     & 0.422                      & 0.415                      & 0.427                      & 0.412   \\ \bottomrule[1pt]
\end{tabular}
}
\caption{Ablation study of various loss terms in the learning objective for the distinguishability loss.}
\label{tab:ablation_experiment}
\vspace{-0.5cm}
\end{table}

\subsection{Ablation Study}
To study the effect of different components of our proposed distinguishability loss, i.e., the constraints applied to the transformation operation for ratio balancing, we remove one of them and compare the performance changes in Table~\ref{tab:ablation_experiment}. Here, $\mathcal{L}_{orth}$ is applied on $W$ in Eq.\ref{eq:orth}, $\mathcal{L}_t$ is applied on \yhq{$S_k$} (from Eq.\ref{eq:lt} and Eq.\ref{eq:lt2}), and $l_2$ is the weight for the $L_2$ regularisation term on all the other learnable parameters.
The $\mathcal{L}_{orth}$ and $L_2$ constraints have similar effects on the overall performance, as they both act as axis transformations, \yhq{while the constraint $L_{t}$ applied on $S_k$ }plays a more important role, whose removal leads to a larger performance drop among all the settings. It partly demonstrates the importance of the balancing ratio vector after the rotation transformation.

\section{Conclusion}
In this paper, to address the information diffusion issue in prompt-based few-shot learning, we propose a calibration method based on featuretransformation which first rotates output embeddings into a new metric space, and then scales the ratio of each dimension to a uniform distribution to guarantee the distinguishability of the transformed embeddings. On the other hand, we utilise hyperbolic embeddings to capture the hierarchical relations between class labels to guide the metric learning strategy to enhance the interpretability of the learned output embeddings. Extensive experiments on the three multi-class classification tasks under various settings demonstrate the effectiveness of our approach with an average {5.9\%} performance improvement on the F1-measure.

\section*{Limitation}

In this work, we only focus on the multi-class classification task with hierarchical class labels. 
Future work could explore extending our idea to other tasks, such as controllable text generation, which has the similar information diffusion issue. Another potential direction in future work is to learn a prior distribution rather than simply using the uniform distribution in ratio balancing. Since the uniform distribution-based ratio balancing is a strong assumption, it might not be suitable for some tasks in real-world applications. One could use VAE or VQ-VAE to learn a distribution which could be subsequently used to regularise the optimisation of feature transformation.

\section*{Acknowledgements}
This work was supported in part by the UK Engineering and Physical Sciences Research Council (EP/T017112/1, EP/V048597/1, EP/X019063/1), and the National Science Foundation (NSF) grant 1750978. Yulan He is supported by a Turing AI Fellowship funded by the UK Research and Innovation (EP/V020579/1).


\bibliography{anthology,custom}
\bibliographystyle{acl_natbib}
\clearpage
\appendix

\setcounter{table}{0}
\renewcommand{\thetable}{A\arabic{table}}
\setcounter{figure}{0}
\renewcommand{\thefigure}{A\arabic{figure}}

\section{Appendix}
\label{sec:appendix}

\subsection{Model Selection}
Following previous research \cite{DBLP:conf/acl/GaoFC20,DBLP:conf/acl/HambardzumyanKM20,DBLP:conf/emnlp/LesterAC21}, BERT \cite{DBLP:conf/naacl/DevlinCLT19}, Roberta and ALBERT \cite{DBLP:conf/iclr/LanCGGSS20} were used when using the \emph{cloze} prompts. The \emph{cloze} is to fill in the blanks in the prompt template by the model itself.

\begin{table}[htbp]
\begin{center}
\resizebox{0.45\textwidth}{!}{%
\begin{tabular}{lccc}
\toprule[1pt]
\textbf{Model}                & \multicolumn{3}{c}{\textbf{Zero-shot}}     \\ \hline
                     & \textbf{Accuracy} & \textbf{Macro-f1} & \textbf{Weighted-f1} \\ \hline
\textbf{\begin{tabular}[c]{@{}l@{}}BERT\\ (fine-tuning)\end{tabular}}         & 0.0342  &0.0196 &  0.0329 \\
\hline
\textbf{BERT}           & 0.0716   & 0.0165   & 0.0384      \\ \hline
\textbf{RoBERTa}     & 0.1094  & 0.0465   & 0.0994     \\ \hline

\textbf{ALBERT} & 0.0538   & 0.0217   &0.0459      \\  
\bottomrule[1pt]
\end{tabular}
}
\end{center}
\caption{Classification results on GoEmotion dataset of different baseline models.}
\label{tab:model_res}
\end{table}

To select the baseline model used as the backbones of our proposed method, we evaluate the baseline models on \textsl{GoEmotions} dataset for zero-shot learning, the results are shown in Table \ref{tab:model_res}. We compare the effects of the same model using prompt learning and fine-tuning, respectively (difference in effects between BERT \cite{DBLP:conf/naacl/DevlinCLT19} and BERT (fine-tuning). After comparison, we chose RoBERTa \cite{liu2019roberta} as it shows the overall best performance.

Based on the large pretrained language model  backbone, we compare different prompt-based training methods and select the best as our \textsl{Prompt-Baseline}. The details are shown in Table~\ref{tab:template_res}.

\subsection{Weight Initialisation}
\label{sec:weight_initial}
The optimisation of $W_k$ and $S_k$ can be affected by different weight initialisations. As such, we experiment with different initialisation strategies and show the results of 100-shot learning in Table~\ref{tab:initialisation} (We use the same initialisation for $W_k$ and $S_k$.). The Gaussian distribution initialisation performs the best overall. Therefore we use the Gaussian distribution initialisation in all the experiments reported in the paper.

\begin{table}[htbp]
\centering
\resizebox{0.45\textwidth}{!}{%
\begin{tabular}{lccc}
\toprule[1pt]
\textbf{Initialisation}   & \textbf{Emotion} & \textbf{WOS} & \textbf{GoEmotions}   \\\midrule[1pt]
\textbf{Gaussian}    & 0.802    & 0.757  & 0.448       \\ 
\textbf{Xavier}      & 0.817    & 0.749  & 0.420       \\
\textbf{Eye}         & 0.798    & 0.747  & 0.387       \\
\textbf{Orthogonal}  & 0.801    & 0.757  &  0.431       \\
\bottomrule[1pt]
\end{tabular}
}
\caption{Initialisation of different distributions on weight matrix.}
\label{tab:initialisation}
\end{table}

\begingroup
\setlength{\tabcolsep}{6pt} 
\renewcommand{\arraystretch}{1} 

\begin{table*}[t!]
\centering
\begin{tabular}{l|ccc|ccc|ccc}
\toprule[1pt]
\multicolumn{1}{l|}{} & \multicolumn{3}{c|}{\textbf{Emotion}}          & \multicolumn{3}{c|}{\textbf{WOS}}             & \multicolumn{3}{c}{\textbf{GoEmotions}}  \\ \hline
\textbf{$K$-shot}             & \textbf{Soft}  & \textbf{P-Tuning}       & \textbf{PTR}            & \textbf{Soft}  & \textbf{P-Tuning}       & \textbf{PTR}            & \textbf{Soft}  & \textbf{P-Tuning} & \textbf{PTR}   \\ \hline
\textbf{5}             & 0.295 & \textbf{0.336} & 0.330          & 0.165 & \textbf{0.236}    & 0.213   & 0.072 & 0.135          & \textbf{0.161} \\ \hline
\textbf{10}   & 0.312 & \textbf{0.363} & 0.351          & 0.180 & \textbf{0.252}    & 0.230 & 0.151 & 0.151          & \textbf{0.173} \\ \hline
\textbf{50}     & 0.363 & \textbf{0.431} & 0.409          & 0.328 & \textbf{0.359}    & 0.319 & 0.230 & 0.245          & \textbf{0.281}\\ \hline
\textbf{100}    & 0.423 & 0.625          & \textbf{0.631} & 0.412 & \textbf{0.435}    & 0.391 & 0.331 & \textbf{0.336} & 0.310 \\ 
\bottomrule[1pt]
\end{tabular}
\caption{Weighted F1 of few-shot for different prompt-based training methods.}
\label{tab:template_res}
\end{table*}

\section{Visualisation results}
To better compare the results of Baseline methods and ours, we visualize the output of different labels by mapping them into 2D plane via T-SNE (Figure~\ref{app:tsne_fig}). It is clear that our model separates the data points of different labels ((b) and (d)) rather than mixing them up (shown in (a) and (c)). To explore the corresponding effects of singular values distribution, we visualise the normalized singular value distribution of the output embeddings in Figure~\ref{fig:sv_dis}. We observe a more balanced distribution after applying our transformation and metric learning.
\begin{figure*}[t]
\centering
\subfigure[P-Tuning on Emotion]{
\begin{minipage}[t]{0.23\linewidth}
\centering
\includegraphics[width=\textwidth,trim={0 0 150 0},clip]{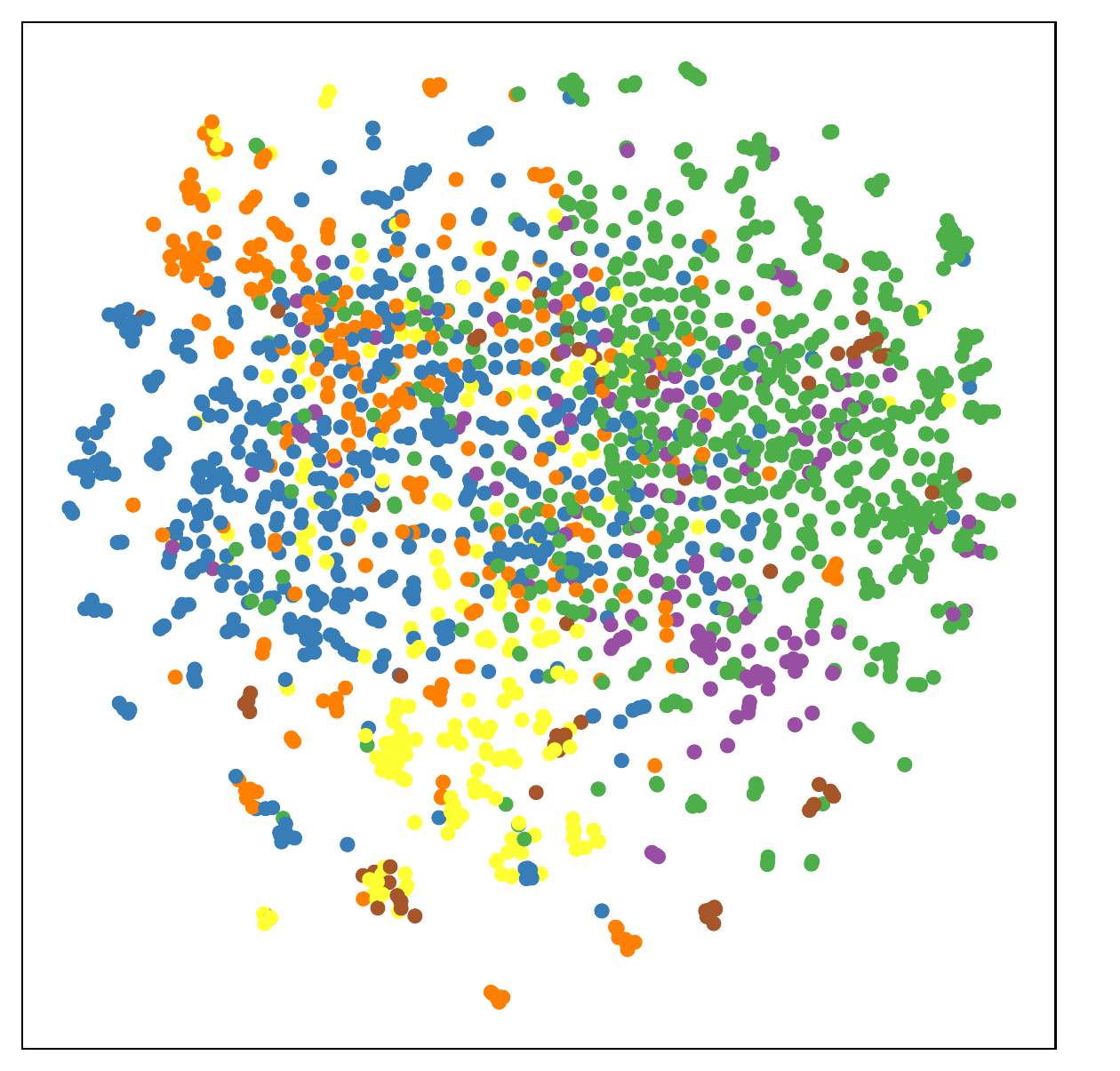}
\end{minipage}%
}%
\subfigure[Our Model on Emotion]{
\begin{minipage}[t]{0.23\linewidth}
\centering
\includegraphics[width=\textwidth,trim={0 0 150 0},clip]{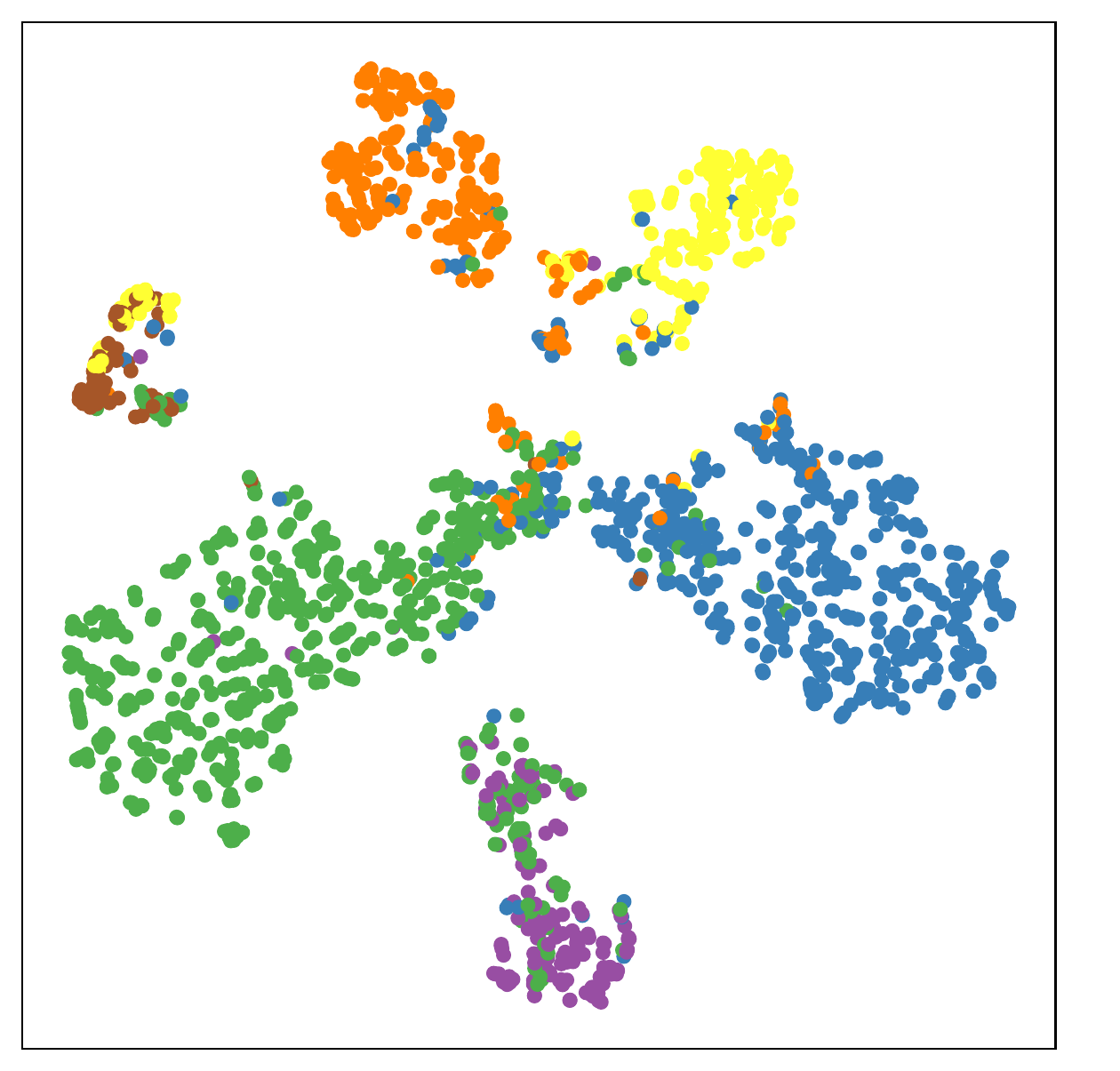}
\end{minipage}%
}%
\centering
\subfigure[PTR on GoEmotion]{
\begin{minipage}[t]{0.23\textwidth}
\centering
\includegraphics[width=\linewidth,trim={0 0 460 0},clip]{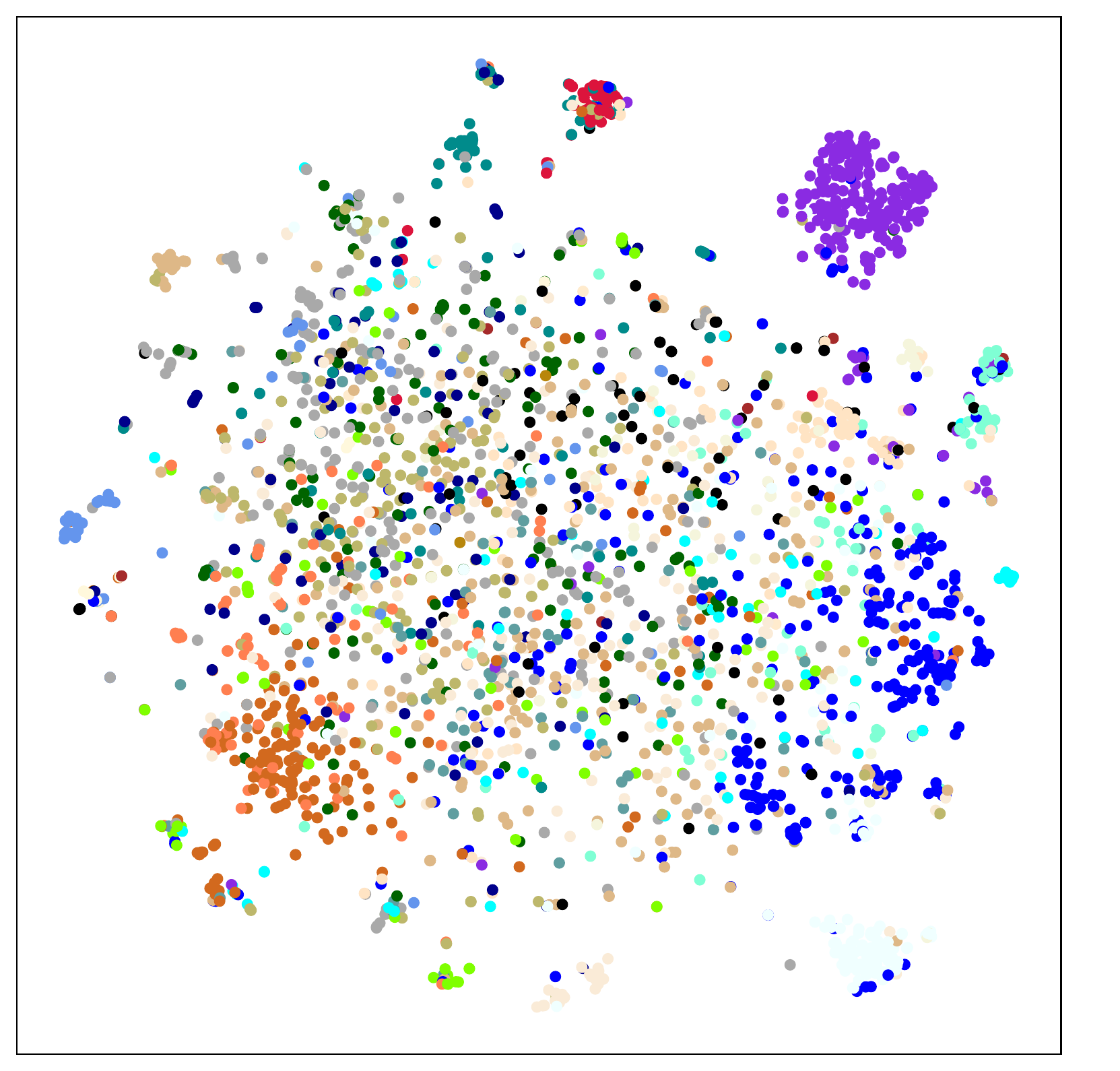}
\end{minipage}%
}
\centering
\subfigure[Our Model on GoEmotion]{
\begin{minipage}[t]{0.23\textwidth}
\centering
\includegraphics[width=\linewidth,trim={0 0 460 0},clip]{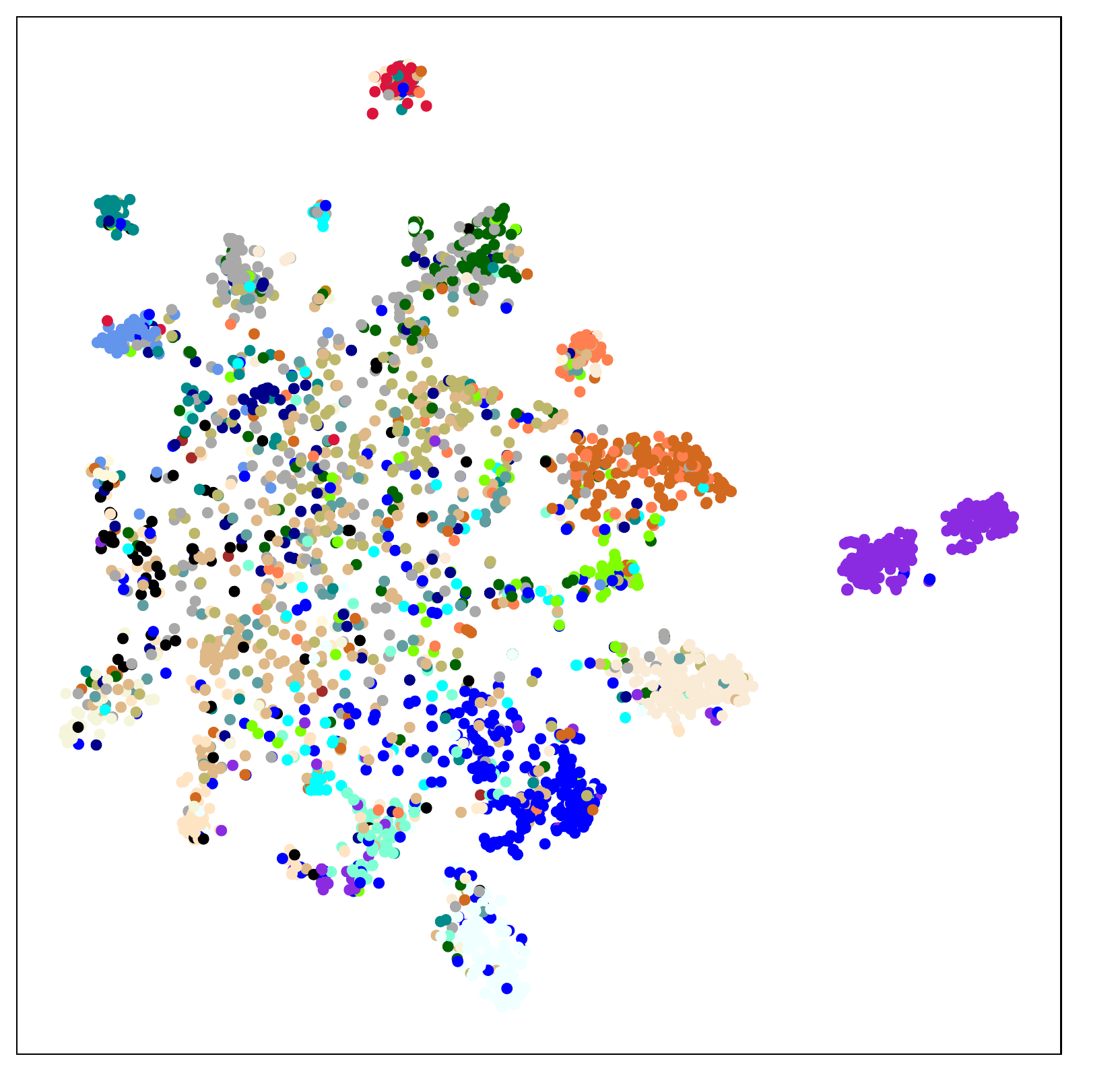}
\end{minipage}%
}
\caption{The T-SNE results of test samples in Emotion/GoEmotions Dataset under 100-shot.}
\label{app:tsne_fig}
\end{figure*}
\begin{figure*}[htbp]
\centering
\subfigure[P-Tuning on Emotion dataset]{
\begin{minipage}[t]{0.45\textwidth}
\centering
\includegraphics[width=\linewidth]{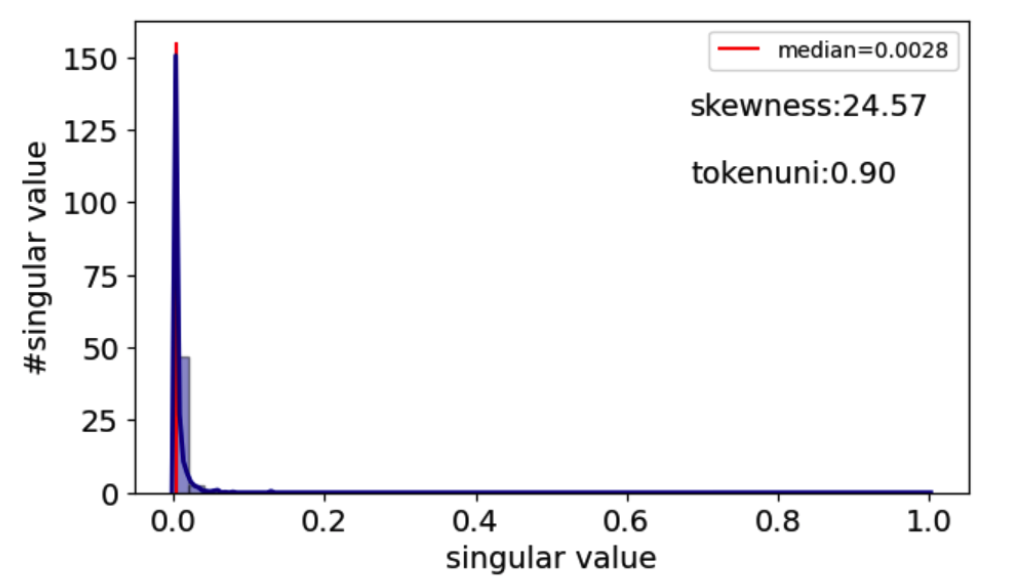}
\end{minipage}%
}%
\subfigure[Our Model on Emotion dataset]{
\begin{minipage}[t]{0.45\textwidth}
\centering
\includegraphics[width=\textwidth]{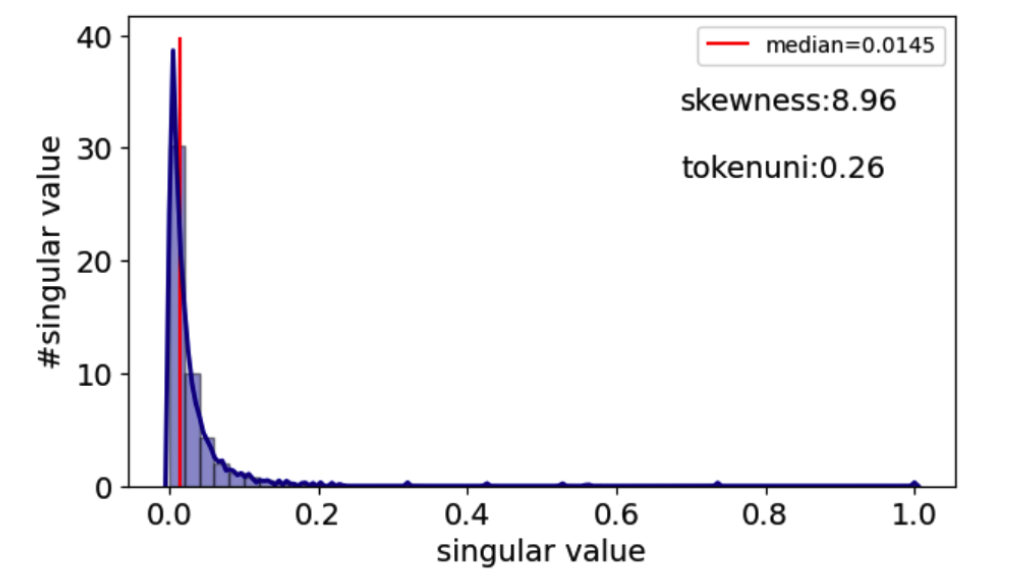}
\end{minipage}%
}
\subfigure[PTR on GoEmotion dataset]{
\begin{minipage}[t]{0.45\textwidth}
\centering
\includegraphics[width=\textwidth]{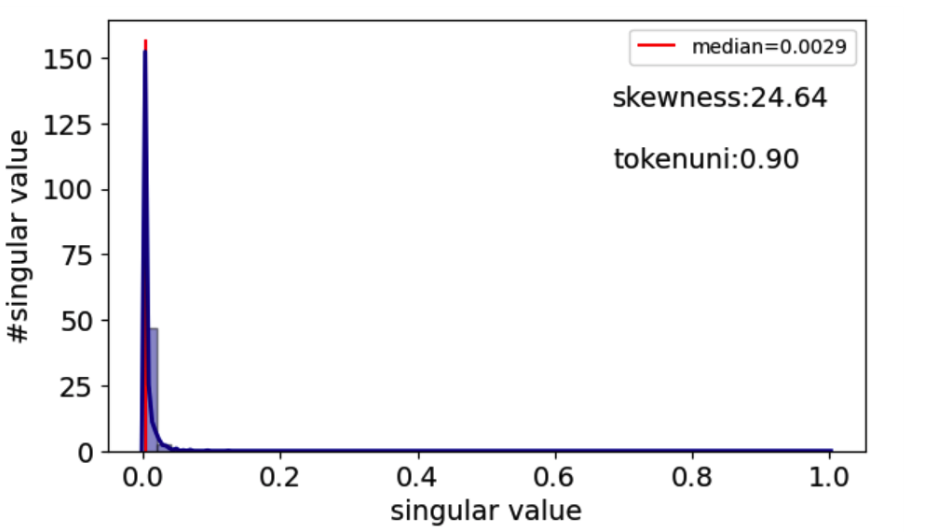}
\end{minipage}%
}%
\subfigure[Our Model on GoEmotion dataset]{
\begin{minipage}[t]{0.36\textwidth}
\centering
\includegraphics[width=\textwidth]{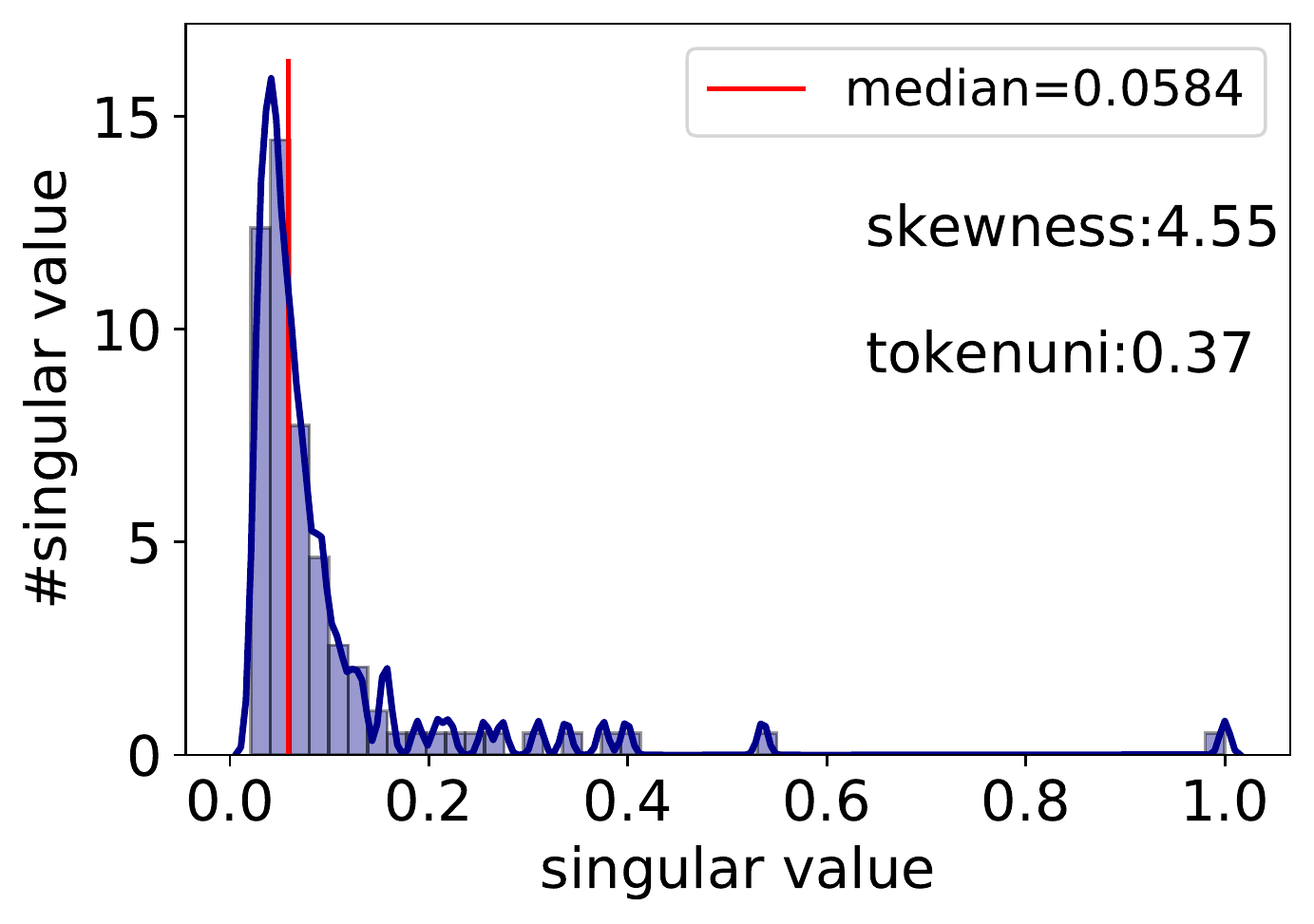}
\end{minipage}%
}%
\caption{The singular value distribution of test samples under 100-shot. Our methods greatly balance the singular distribution, i.e., decrease the skewness, and alleviate the \emph{information diffusion} issue, i.e., decrease the token similarity (\emph{tokenuni}).}
\label{fig:sv_dis}
\end{figure*}

\end{document}